\documentclass[11pt,a4paper,nohyperref]{article}
\pdfoutput=1
 
\usepackage{times,latexsym}
\usepackage{url}
\usepackage[T1]{fontenc}

\usepackage[utf8]{inputenc}
\usepackage{xcolor}
\usepackage{amsfonts}       
\usepackage{nicefrac}       
\usepackage{microtype}      
\usepackage{amsmath}
\usepackage{amssymb}
\usepackage{amsthm}
\usepackage{booktabs}
\usepackage{color}
\usepackage{graphicx}
\usepackage{multirow}
\usepackage{subfigure}
\usepackage{url}
\usepackage{verbatim}
\usepackage{tipa}

\definecolor{gray}{gray}{0.5}

\usepackage[acceptedWithA]{tacl2018v2}

\usepackage{xspace,mfirstuc,tabulary}

\newif\iftaclinstructions
\taclinstructionsfalse 
\iftaclinstructions
\renewcommand{\confidential}{}
\renewcommand{\anonsubtext}{(No author info supplied here, for consistency with
TACL-submission anonymization requirements)}
\newcommand{\instr}
\fi

\iftaclpubformat 

\else

\fi

\title{Learning Robust and Multilingual Speech Representations}

\author{Kazuya Kawakami$^{\clubsuit\spadesuit}$
        Luyu Wang$^{\clubsuit}$
        Chris Dyer$^{\clubsuit}$
        Phil Blunsom$^{\clubsuit\spadesuit}$
        Aaron van den Oord$^{\clubsuit}$\\
$^{\clubsuit}$DeepMind, London, UK\\
$^{\spadesuit}$Department of Computer Science, University of Oxford, Oxford, UK \\\
{\small \tt \{kawakamik,luyuwang,cdyer,pblunsom,avdnoord\}@google.com}
}

\begin{document}
\maketitle
\begin{abstract}
Unsupervised speech representation learning has shown remarkable success at finding representations that correlate with phonetic structures and improve downstream speech recognition performance. However, most research has been focused on evaluating the representations in terms of their ability to improve the performance of speech recognition systems on read English (e.g. Wall Street Journal and LibriSpeech). This evaluation methodology overlooks two important desiderata that speech representations should have: robustness to domain shifts and transferability to other languages. In this paper we learn representations from up to 8000 hours of diverse and noisy speech data and evaluate the representations by looking at their robustness to domain shifts and their ability to improve recognition performance in many languages. We find that our representations confer significant robustness advantages to the resulting recognition systems: we see significant improvements in out-of-domain transfer relative to baseline feature sets and the features likewise provide improvements in 25 phonetically diverse languages including tonal languages and low-resource languages.   
\end{abstract}

\section{Introduction}\label{sec:intro}
The input representation of machine learning model strongly determines the difficulty faced by the learning algorithm, how much data the learner will require to find a good solution, and whether the learner generalizes out of sample and out of the domain of the training data. Representations (or features) that encode relevant information about data enable models to achieve good performance on downstream tasks, while representations that are invariant to factors that are not relevant to downstream tasks can further improve generalization. Traditionally, many invariances were hard-coded in feature extraction methods. For example, in image representations, geometric and photometric invariance has been investigated~\citep{mundy1992geometric,van2005robust}. For acoustic representations, it is known that the standard MFCC features are sensitive to additive noise and many modifications have been proposed to overcome those limitations~\citep{dev2010robust,kumar2011delta}.

Recently, unsupervised representation learning algorithms have shown significant improvements at learning representations that correlate well with phonetic structure~\cite{oord2018representation,kahn2019libri} and improving downstream speech recognition performance~\cite{schneider2019wav2vec,baevski2019vq}. Most of this work focused on learning representations from read English speech (from the LibriSpeech and LibriVox datasets) and evaluating the features when used to recognize speech in a rather similar domain (read English text). However, this approach to evaluation fails to test for the invariances that we would like good speech representations to have: robustness to domain shifts and transferability to other languages.

In our experiments we learn representations from 8000 hours of diverse and noisy speech, using an extended version of contrastive predictive coding model: bidirectional predictive models with dense residual connections ~(\S\ref{sec:cpc}--\S\ref{sec:experiments}), and evaluate the robustness and transferability of our representations by estimating how invariant they are to domain and language shifts. To do so, an ASR model is trained using our representations on one dataset but evaluated on the test sets of other datasets. In this experiment, we find that the representations derived from the large pretraining dataset lead the ASR model to be much more robust to domain shifts, compared to both log filterbank features as well as to pretraining just on LibriSpeech. We also train ASR models on 25 languages, including low-resource languages (e.g.\ Amharic, Fongbe, Swahili, Wolof), and show that our representations significantly outperform both standard features and those pretrained only on clean English data in the language transfer setup.

In summary, we confirm several increasingly common patterns that may be discerned in the literature on unsupervised representation learning, across a variety of modalities. First, scale matters: good representation learning requires a large amount of data. Second, unsupervised representations consistently improve robustness on downstream tasks. And finally, representations learned from multilingual data can transfer across many languages.

\section{Contrastive Predictive Coding: CPC}\label{sec:cpc}
Unsupervised representation learning methods rely on differentiable objectives which quantify the degree to which representations have succeeded at capturing the relevant characteristics in data. Mutual information measures relationships between random variables~\citep{fano1961transmission}. Mutual information maximization techniques, that learn representations that describe data by maximizing mutual information between data and representation variables, have been explored for a long time in unsupervised representation learning~\citep{linsker1989application,bell1995information}. However, since the exact computation of mutual information is not tractable for continuous variables, recently many estimators have been proposed for enabling unsupervised representation learning with neural networks~\citep{belghazi2018mine,oord2018representation,poole2019variational}.

Contrastive predictive coding~\citep[CPC]{oord2018representation} is a mutual information maximization method that has been successfully applied to many modalities such as images and speech~\citep{henaff2019data,schneider2019wav2vec}. The objective is designed to extract features that allow the model to make long-term predictions about future observations. This is done by maximizing the mutual information of these features with those extracted from future timesteps. The intuition is that the representations capture different levels of structure dependent on how far ahead the model predicts. For example, if the model only predicts a few steps ahead, the resulting representations can capture local structures. On the other hand, if the model predicts further in the future, the representations will need to infer ``slow features''~\citep{wiskott2002slow}; more global structures such as phonemes, words and utterances in speech.

The overall unsupervised learning process is visualized in Figure \ref{fig:method}. Given a raw audio signal of length $L$ ($\boldsymbol{x} = x_{1}, x_{2},\ldots, x_{L}$, $x_i \in \mathbb{R}$ where $x_i$ represents the acoustic amplitude at time $i$), a function $g_{\textit{enc}}$ encodes the audio signals into vector representations ($\boldsymbol{z} = \boldsymbol{z}_{1}, \boldsymbol{z}_{2}\ldots, \boldsymbol{z}_{M}$, $\boldsymbol{z}\in \mathbb{R}^{d_{z}}$). Next, an autoregressive function $g_{\textit{ar}}$, such as a recurrent neural network, summarizes the past representations and produces context vectors ($\boldsymbol{c} = \boldsymbol{c}_{1}, \boldsymbol{c}_{2}\ldots, \boldsymbol{c}_{M}, \boldsymbol{c}\in\mathbb{R}^{d_{c}}$). The representations are learned to maximize mutual information between context vectors ($\boldsymbol{c}_{t}$) and future latent representations ($\boldsymbol{z+k}$) as follows:
$$ I(\boldsymbol{c}_{t},\boldsymbol{z}_{t+k})= \sum_{\boldsymbol{c}_{t},\boldsymbol{z}_{t+k}} p(\boldsymbol{c}_{t}, \boldsymbol{z}_{t+k}\mid k)\log\frac{p(\boldsymbol{z}_{t+k}\mid \boldsymbol{c}_{t}, k)}{p(\boldsymbol{z}_{t+k})}.
$$

Since the mutual information is not tractable for high dimensional data, it is common to use a lower-bound on the mutual information such as InfoNCE~\citep{oord2018representation} which is a loss function based on noise contrastive estimation~\citep{gutmann2010noise}.
Given a set $Z = \{\boldsymbol{z}_{1},\ldots  \boldsymbol{z}_{N}\}$ which contains one positive sample from $p(\boldsymbol{z}_{t+k}|\boldsymbol{c}_{t})$ and $N-1$ negative samples from a ``noise'' distribution $p(\boldsymbol{z})$, the approximated lower-bound is written as: 
$$ \label{eq:infonce}
    I(\boldsymbol{c}_{t},\boldsymbol{z}_{t+k}) \geq \mathbb{E}_{Z}\left[\log\frac{f_{k}(\boldsymbol{c}_{t}, \boldsymbol{z}_{t+k})}{\frac{1}{N}\sum_{\tilde{\boldsymbol{z}} \in Z}f_{k}(\boldsymbol{c}_{t}, \tilde{\boldsymbol{z}})} \right] = \mathcal{L}^{\textit{NCE}}_{tk},
$$

where $f_{k}(\boldsymbol{c}_{t}, \boldsymbol{z}_{t+k})$ is a scoring function. We used the standard log-bilinear model as follows:
$$
    f_{k}(\boldsymbol{c}_{t}, \boldsymbol{z}_{t+k}) = \exp(\boldsymbol{c}^{T}_{t}\boldsymbol{W}_{k}\boldsymbol{z}_{t+k}).
$$

The loss function we maximize is a sum of the InfoNCE loss for each step, $\mathcal{L}^{\textit{NCE}} = \sum_{t}\sum_{k}\mathcal{L}^{\textit{NCE}}_{tk}$ and the negatives are uniformly sampled from representations in the same audio signal ($\boldsymbol{z}$).

\section{Methods}\label{sec:methods}
\begin{figure*}[htbp]
    \begin{center}
        \includegraphics[width=0.95\linewidth]{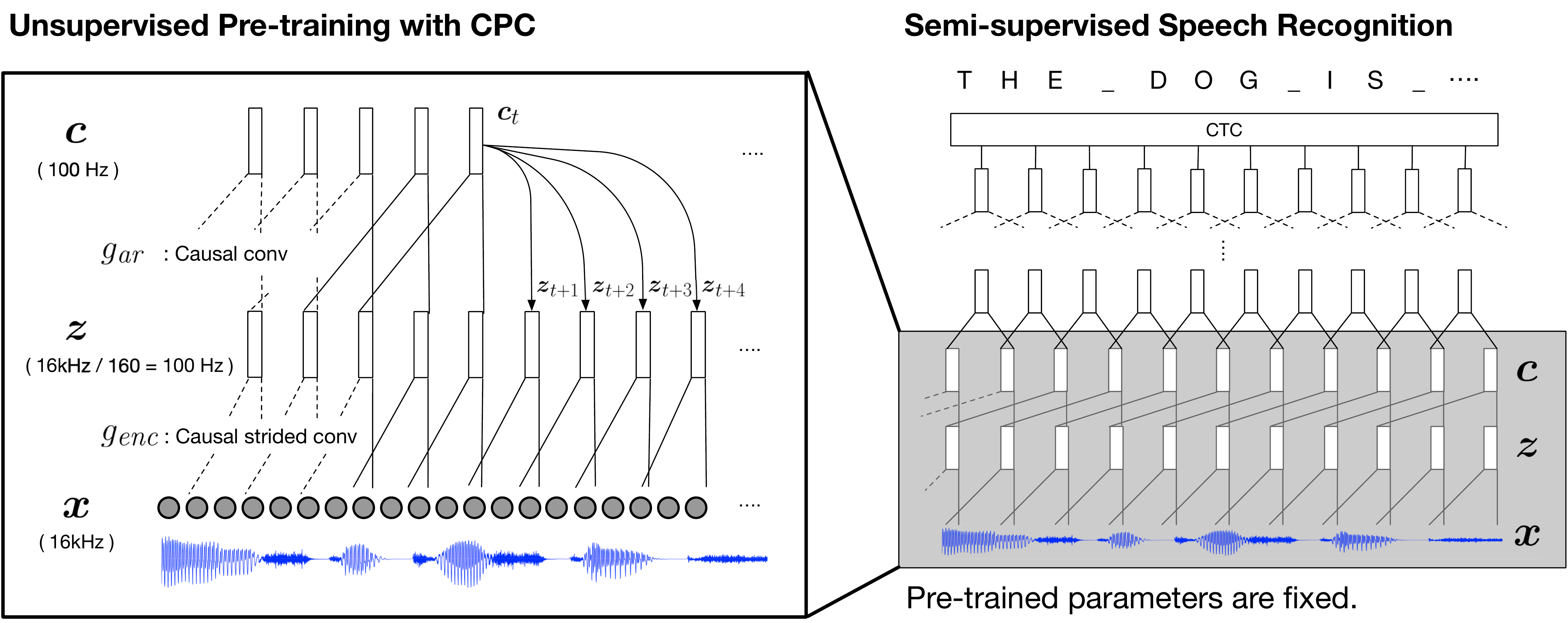}
    \end{center}
    \caption{\textbf{Left}, unsupervised representation learning with forward contrastive predictive coding. The learned representations are fixed and used as inputs to a speech recognition model (\textbf{Right}).}
    \label{fig:method}
\end{figure*}

In this section, we describe our models and objectives for unsupervised representation learning and downstream speech recognition. First, an acoustic feature extractor is trained with a bidirectional variant of contrastive predictive coding on an unlabeled audio dataset. Next, the parameters of this model are frozen and its output representations are used as input to train various speech recognition models, potentially on a different or smaller labeled dataset (Figure \ref{fig:method}).

\subsection{Unsupervised learning with bi-directional CPC}\label{sec:unsup}
Following the success of bidirectional models in representation learning~\citep{Peters:2018,devlin2018bert}, we extend the original CPC method explained above with bidirectional context networks. The encoder function $g_{\textit{enc}}$ is shared for both directions, but there are two autoregressive models ($g^{\textit{fwd}}_{\textit{ar}}$ and $g^{\textit{bwd}}_{\textit{ar}}$) which read encoded observations ($\boldsymbol{z}$) from the forward and backward contexts, respectively. The forward and backward context representations $\boldsymbol{c}^{\textit{fwd}}_{t}, \boldsymbol{c}^{\textit{bwd}}_{t}$ are learned with separate InfoNCE losses. When they are used for downstream tasks, a concatenation of two representations $\boldsymbol{{c}_{t}} = [\boldsymbol{c}^{\textit{fwd}}_{t}; \boldsymbol{c}^{\textit{bwd}}_{t}]$ is used. A similar technique has been used in image representation learning where representations are learned along different spatial dimensions~\citep{henaff2019data}.

All audio signals have a sampling rate of 16kHz and we normalize the mean and variance of the input signals over each utterance in order to mitigate volume differences between samples. For architectures, we use encoder and autoregressive models similar to \cite{schneider2019wav2vec}. The encoder function $g_{\textit{enc}}$, is a stack of causal convolutions with kernel sizes (10, 8, 4, 4, 4, 1, 1) and stride sizes (5, 4, 2, 2, 2, 1, 1), corresponding to a receptive field of 10 ms of audio. For autoregressive functions, we use a 13 layer causal convolution architecture with kernel sizes (1, 2, \ldots, 12, 13) and stride size 1, for both forward and backward functions. Layer-normalization across the temporal and feature dimensions is applied to every layer. Also, each layer has dense skip connections with layers below as in DenseNet~\citep{huang2017densely}. The objective function we optimize is the sum of the forward and backward InfoNCE losses (eq.\ref{eq:infonce}).

\subsection{Semi-supervised speech recognition}\label{sec:ssl}
Once the acoustic representations are trained, the resulting context vectors ($\boldsymbol{c}$) are used as inputs to character-level speech recognition models which predict transcriptions of audio-signals character by character. The model first predicts frame-level character probabilities with a series of convolution layers while the CTC forward algorithm~\citep{graves2006connectionist} calculates conditional probabilities of a transcription given an audio signal. The model parameters are trained to maximize the log likelihood of the data. The training terminates when the word error rate on the development set stops improving or the model has trained for more than a certain number of epochs. The models are evaluated on the standard word error rate (WER) metric on held-out test data. During training, the parameters in the speech recognition models are trained with supervision but the parameters of the pretrained representation models remain fixed. For decoding, we use greedy CTC decoding. In most experiments, we do not use a language model (LM) in order to isolate the effects of the acoustic representations on performance, but we do include results with a 4-gram LM to facilitate comparisons with published results.

Common practice in unsupervised representation learning is to evaluate learned representations using a linear classifier rather than a more complex nonlinear model. However, we find that a simple linear layer followed by a CTC decoder does not have enough capacity to recognize speech. Thus, for our first set of experiments we use a smaller version of DeepSpeech2~\citep{amodei2016deep} to predict the frame-level character probabilities. The model has two 2d-convolutions with kernel sizes (11, 41) and (11, 21) and stride sizes (2, 2) and (1, 2) and one unidirectional recurrent neural network (GRU) on top of the output from the convolution layers. A linear transformation and a softmax function are applied to predict frame-level character probabilities. We refer to \textbf{DeepSpeech2 small} for the model specifics~\citep{amodei2016deep}. In order to further investigate how the representations interact with larger speech recognition models, we use the time-delay neural networks (\textbf{TDNN}) that are commonly used in speech recognition~\citep{collobert2016wav2letter,openseq2seq}. These consist of 17 layers of 1d-convolutions followed by 2 fully connected layers. Refer to OpenSeq2Seq for a detailed description.\footnote{\url{https://nvidia.github.io/OpenSeq2Seq/html/speech-recognition/wave2letter.html}} These large models have been designed to perform well with log-filterbank features and purely supervised learning on large datasets, so they represent a challenging and informative test case for the value of learned representations.

\section{Experiments and Results}\label{sec:experiments}

\subsection{Datasets}\label{sec:dataset}
We collected publicly available speech datasets which cover a variety of types of speech (e.g.\ read and spoken), noise conditions and languages. For unsupervised pretraining we use a combination of datasets, using the audio but not any transcriptions, even when they are available. For semi-supervised learning (i.e., evaluation) on top of the representations we use the transcribed datasets following their standard train-test splits. Table \ref{tb:dataset} summarizes the datasets used for unsupervised learning and English speech recognition tasks.

\begin{table}[ht]
    \begin{center}\small
    \begin{tabular}{lrrr}
    \toprule
    Name  & \multicolumn{1}{c}{Language} & \multicolumn{1}{c}{Type} & \multicolumn{1}{c}{Hours}\\
    \midrule
    Audio Set & Multilingual & - & 2500\\ 
    AVSpeech & Multilingual & - & 3100\\ 
    Common Voice & Multilingual & read & 430\\
    \midrule
    LibriSpeech & English & read & 960\\
    WSJ & English & read & 80\\     
    TIMIT & English & read & 5\\     
    SSA & English & read & $<$1\\ 
    \midrule
    Tedlium & English & spoken & 440\\
    Switchboard & English & spoken & 310\\ 
    \bottomrule
    \end{tabular}
    \end{center}
    \caption{Summary of English Datasets.\label{tb:dataset}}
\end{table}

\paragraph{Unlabeled speech pretraining corpus}
For pretraining, we collected a diverse and noisy speech corpus from several existing datasets: the subset of Audio Set~\citep{jort2017audioset} containing speech examples, the audio part of AVSpeech~\citep{ephrat2018looking}, and the Common Voice (CV)\footnote{\url{https://voice.mozilla.org}} dataset in all 29 available languages. In addition we used the audio from TIMIT~\citep{garofolo1993timit} and the Speech Accent Archive~\citep{weinberger2009towards}, ignoring the transcriptions. Finally, we include the audio (again ignoring transcriptions) from the standard training splits of the evaluation datasets below. This collection spans a range of recording conditions, noise levels, speaking styles, and languages and amounts to about 8000 hours of audio.

\paragraph{Transcribed read English}
For evaluation, we look at the performance of our representations on a variety of standard English recognition tasks, as well as their ability to be trained on one and tested on another. For read English, we use LibriSpeech~\citep{panayotov2015librispeech} and the Wall Street Journal~\citep{paul1992design}.

\paragraph{Transcribed spoken English}
To explore more extreme domain shifts, we additionally used conversational speech and public speaking datasets. We used Switchboard~\citep{godfrey1992switchboard}, a standard conversational speech recognition dataset consisting of two-sided telephone conversations (test only). Since the data was recorded more than 10 years ago and at a lower sampling rate than the other corpora, it presents a noisy and challenging recognition problem. Finally, we also use the Tedlium-3~\citep{hernandez2018ted} corpus, a large spoken English dataset containing 450 hours of speech extracted from TED conference talks. The recordings are clear, but there is some reverberation.

\paragraph{Transcription normalization}
Since we are comparing ASR systems trained on one dataset but evaluated on the test set of another, we normalize transcriptions to avoid systematic biases in the transfer condition. To do so, we use the format of the LibriSpeech dataset, which also ensures that our results are comparable with standard speech recognition systems on that task~\citep{openseq2seq}. For the remaining datasets, the transcriptions are lower-cased and unpronounced symbols (e.g., punctuation, silence markers, etc.) are removed. We also remove utterances that contain numbers as they are transcribed inconsistently across and within datasets.

\paragraph{Transcribed multilingual speech}
In order to evaluate the transferability of the representations, we use speech recognition datasets in 4 African languages collected by the ALFFA project,\footnote{\url{http://alffa.imag.fr}} Amharic~\citep{tachbelie2014}, Fongbe~\citep{laleye2016FongbeASR}, Swahili~\citep{gelas:hal-00954048}, Wolof~\citep{gauthier2016collect}, for evaluation. These languages have unique phonological properties (e.g.~height harmony) and phonetic inventories, making them a good contrast to English. These African languages are low-resource, each with 20 hours or less of transcribed speech. We also use 21 phonetically diverse languages from OpenSLR.\footnote{\url{https://openslr.org}. We only include datasets containing more than 1GB of audio. When there is more than one dataset available for one language, we used the largest dataset.} Table ~\ref{tb:multilingual} summarizes the multilingual dataset statistics.

\subsection{Unsupervised Representation Learning}
We train the model described above~(\S\ref{sec:unsup}) using the datasets described in the previous section~(\S\ref{sec:dataset}).  Similarly to  \citet{schneider2019wav2vec}), audio signals are randomly cropped with a window size 149,600 observations (9.35 seconds) and encoded with the model. The bidirectional contrastive predictive coding objective (Eq.~\ref{eq:infonce}) with prediction steps ($k$) 12 and negatives ($N$) 10 is optimized with the Adam optimizer with learning rate $0.0001$. A batch size of 128 is used as well as a polynomial learning rate scheduler with power 2 and gradient clipping with maximum norm 5.0. Training was terminated at 4.2 million steps based on speech recognition performance on the dev (= validation) set of the LibriSpeech corpus.

\begin{table*}[ht]
    \begin{center}
    \begin{tabular}{lrrrrrrrr}
    \toprule
     & \multicolumn{2}{c}{\textbf{WSJ}} & \multicolumn{2}{c}{\textbf{LibriSpeech}} & \multicolumn{2}{c}{\textbf{Tedlium}} & \multicolumn{1}{c}{\textbf{Switchboard}}\\
     & \multicolumn{1}{c}{test92} & \multicolumn{1}{c}{test93}
     & \multicolumn{1}{c}{test-clean} & \multicolumn{1}{c}{test-other} & \multicolumn{1}{c}{dev} & \multicolumn{1}{c}{test} & \multicolumn{1}{c}{eval2000}\\
    \midrule
    \multicolumn{8}{l}{\textbf{WSJ}}\\
    LogFilterbank & \textcolor{gray}{16.78} & \textcolor{gray}{23.26} & 46.27 & 73.27 & 58.61 & 62.55 & 96.44\\
    CPC-LibriSpeech & \textcolor{gray}{11.89} & \textcolor{gray}{15.66} & 31.05 & 56.31 & 45.42 & 47.79 & 83.08\\
    CPC-8k & \textcolor{gray}{\textbf{10.77}} & \textcolor{gray}{\textbf{14.99}} & \textbf{29.18} & \textbf{51.29} & \textbf{38.46} & \textbf{39.54} & \textbf{69.13}\\
    \midrule
    \multicolumn{8}{l}{\textbf{LibriSpeech}}\\
    LogFilterbank &	14.42 & 21.08 & \textcolor{gray}{6.43} & \textcolor{gray}{20.16} & 26.9 & 25.94 & 61.56\\
    CPC-LibriSpeech & 14.28 & 20.74 & \textcolor{gray}{6.91} & \textcolor{gray}{21.6} & 26.53 & 27.14 & 63.69\\
    CPC-8k & \textbf{13.31} & \textbf{18.88} & \textcolor{gray}{\textbf{6.25}} & \textcolor{gray}{\textbf{19.10}} & \textbf{21.56} & \textbf{21.77} & \textbf{53.02}\\
    \midrule
    \multicolumn{8}{l}{\textbf{Tedlium}}\\
    LogFilterbank & 20.35 & 27.23 & 24.05 & 47.27 & \textcolor{gray}{18.75} & \textcolor{gray}{19.31} & 74.55\\
    CPC-LibriSpeech& 15.01 & 19.52 & 17.77 & 36.7 & \textcolor{gray}{15.28} & \textcolor{gray}{15.87} & 61.94\\
    CPC-8k & \textbf{13.17} & \textbf{17.75} & \textbf{16.03} & \textbf{32.35} & \textcolor{gray}{\textbf{13.67}} & \textcolor{gray}{\textbf{13.88}} & \textbf{47.69}\\
    \bottomrule
    \end{tabular}
    \end{center}
    \caption{Domain transfer experiments to test the robustness of the representations to domain shifts. The models are trained on the \textbf{Wall Street Journal}, \textbf{LibriSpeech} or \textbf{Tedlium} and evaluated on different evaluation sets. The results on in-domain evaluation sets are in gray color. All the results are without a language model. \label{tb:result-transfer}}
\end{table*}

\subsection{Robustness}
Robustness to shifts in domain, recording conditions, and noise levels is an important desideratum for a good ASR system, and we hypothesized that the diversity of our largest pretraining regime would improve robustness along these dimensions. In contrast, standard MFCC features have been tested in terms of noise robustness and it is known that such representations are sensitive to additive noise~\citep{zhao2013analyzing}. Moreover, speech recognition systems developed on top of such features are not robust when they are evaluated on out-of-domain datasets~\citep{amodei2016deep}.

To test whether our pretraining approach improves robustness, we evaluate speech recognition models trained on the learned representations on many different datasets so as to investigate benefit of using the representations learned from large-scale data. We compare ASR systems on all of the Wall Street Journal and LibriSpeech corpora with the same optimization as explained above and evaluate word error rate on different evaluation sets, such as phone call conversations (Switchboard).

Table \ref{tb:result-transfer} summarizes the results on models trained on Wall Street Journal, LibriSpeech or the Tedlium corpora and evaluated on different evaluation sets. \textbf{CPC-LibriSpeech} and \textbf{CPC-8k} indicate representations are learned from LibriSpeech and 8000h of speech datasets listed above respectively. The features trained on large-scale data consistently outperform other representations across different evaluation sets. The speech recognition models trained on the Wall Street Journal perform badly on phone call data in general. However, CPC representations learned on large datasets are more robust than those trained only on read English data (LibriSpeech).

\subsection{Low-resource Languages}
Thus far, all our experiments have compared our representations in terms of their impacts on English recognition tasks (although we know that the pretraining dataset contains samples from many languages). We now turn to the question of whether these representations are suitable for driving recognition different languages with substantially different phonetic properties than English has. Specifically, we look at the performance on four languages---Amharic, Fongbe, Swahili, and Wolof---which manifest a variety of interesting phonological properties that are quite different from English. Evaluating on such languages will provide insights into the phonetic space learned in the representations. Moreover, our non-English languages are low-resource in terms of speech recognition data, but have 2--20 million native speakers each. It is therefore valuable if the representations learned from large-scale unlabelled data can improve low-resource speech recognition. Although there is a small chance that the large-scale pretraining dataset may contain some examples from those languages, we did not add any extra data specifically to improve representations for those languages.

To test the cross-linguistic value of these features, we trained speech recognition models on low-resource languages (\S\ref{sec:dataset}) and compare the relative reduction in WER by switching from standard spectrogram features and the learned representations. As these are very small datasets, we trained \textbf{DeepSpeech2 small} models with the Adam optimizer with a fixed learning rate of $0.0002$ and gradient clipping with maximum norm 25.0. Note that we did not tune architectures and learning methods for particular languages.

\begin{figure}[ht]
    \begin{center}
        \includegraphics[width=1.0\linewidth]{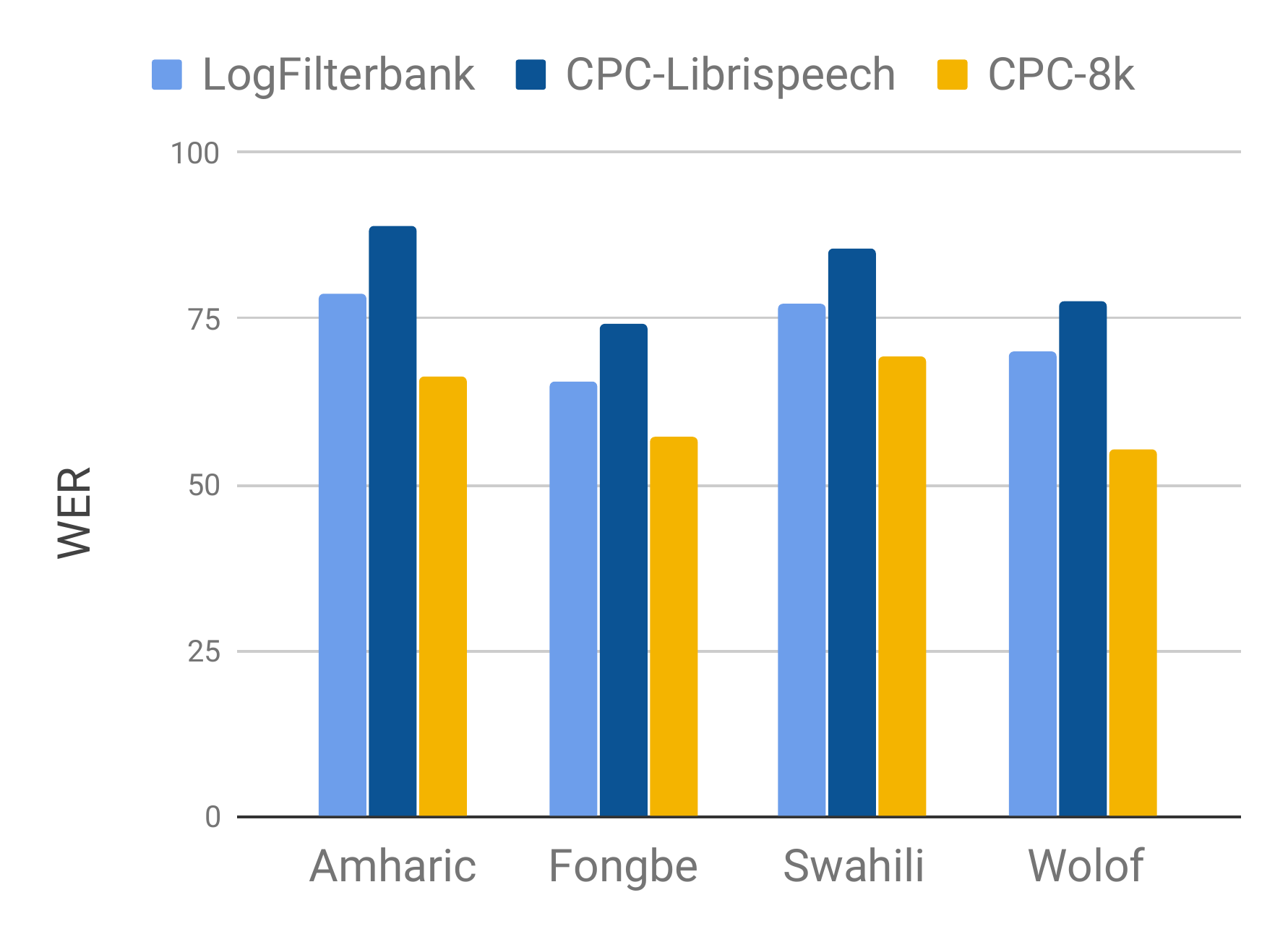}
    \end{center}
    \caption{Speech recognition performance on low-resource African languages (in word error rate). CPC features trained on diverse datasets features significantly outperform baseline log-filterbank features whereas the features trained only on English underperform the baseline.
    \label{fig:lowresource}}
\end{figure}

Figure~\ref{fig:lowresource} summarizes results. Again, we find that the CPC-8k representations outperform other features by a large margin and that the models trained on the representations trained on using the audio of (English-only) LibriSpeech do not perform even as well as basic spectrogram features. This suggests that the representations learned on large-scale data capture a phonetic space that generalizes across different languages, but that diversity of linguistic inputs is crucial for developing this universality.

\begin{table}[ht]
    \begin{center}\small
    \begin{tabular}{l@{\hskip 4pt}rrrrr}
    \toprule
        Language name & Code & Dataset & Hours\\
     \midrule
        Amharic & am & ALFFA & 18.3 \\
        Fongbe & fon & ALFFA & 5.2 \\
        Swahilli & sw  &ALFFA & 8.9 \\
        Wolof & wo & ALFFA & 16.8 \\
     \midrule
        Czech & cs & OpenSLR-6 & 15.0 \\ 
        Uyghur & ug & OpenSLR-22 & 20.2 \\
        Javanese & jv & OpenSLR-35 & 236.8 \\
        Sundanese & su & OpenSLR-36 & 265.9 \\
        Tunisian Arabic & ar & OpenSLR-46 & 4.5 \\
        Sinhala & si & OpenSLR-52 & 179.6 \\
        Bengali & bn & OpenSLR-53 & 172.3 \\
        Nepali & ne	& OpenSLR-54 & 123.6 \\
        African French & fr	& OpenSLR-57 & 13.7 \\
        Catalan & ca & OpenSLR-59 & 71.9 \\
        Malayalam  & ml & OpenSLR-63 & 4.4 \\
        Tamil & ta & OpenSLR-65 & 5.7 \\
        Spanish & es & OpenSLR-67 & 19.6 \\
        Nigerian English & en & OpenSLR-70 & 39.5 \\
        Chilean Spanish & es	& OpenSLR-71 & 5.7 \\
        Columbian Spanish & es & OpenSLR-72 & 6.1 \\
        Peruvian Spanish & es & OpenSLR-73 & 7.3 \\
        Basque & eu & OpenSLR-76 & 11.0 \\
        Galician & gl & OpenSLR-77 & 8.2 \\
        Gujarati & gu & OpenSLR-78 & 6.3 \\
        Kannada & kn & OpenSLR-79 & 6.7 \\
    \bottomrule
    \end{tabular}
    \end{center}
    \caption{Summary of Multilingual Datasets.\label{tb:multilingual}}
\end{table}

\subsection{Multilingual Transfer}
\begin{figure*}[ht]
    \begin{center}
        \includegraphics[width=0.95\linewidth]{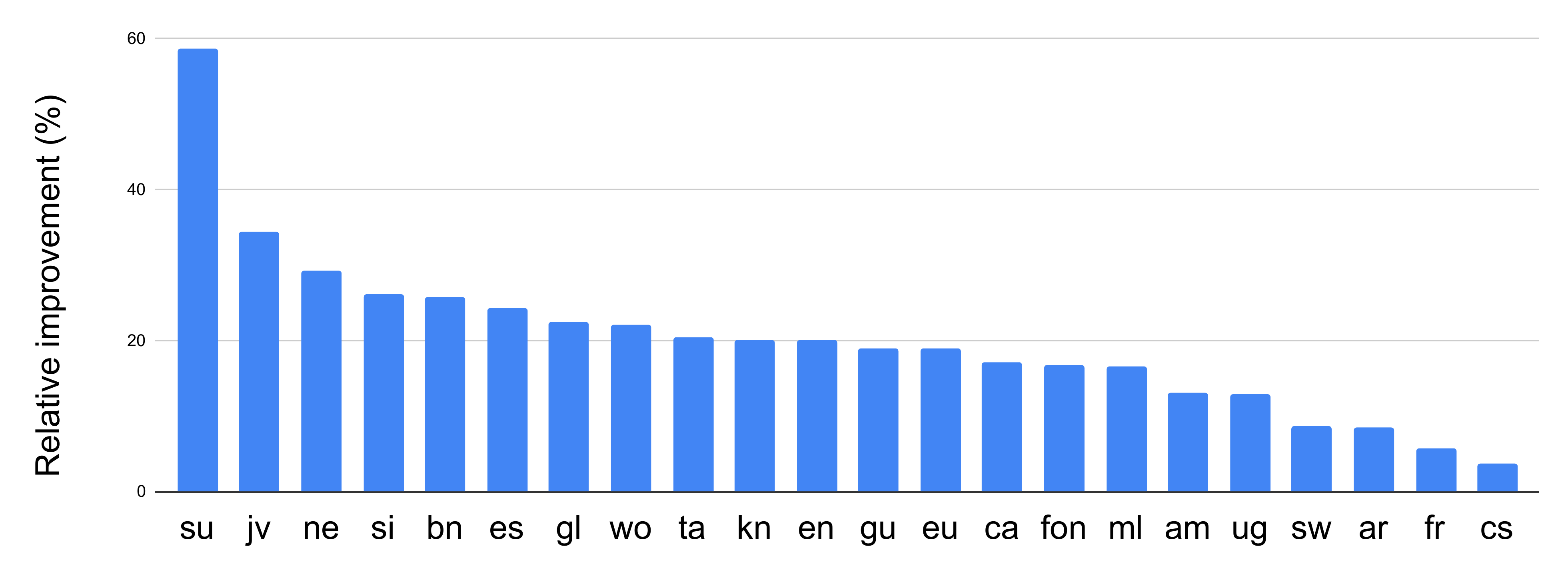}
    \end{center}
    \caption{Relative improvements (in percentage) on speech recognition on many languages with CPC-8k features over Spectrogram features. Each column correspond to language code explained in Table \ref{tb:multilingual}. Note that \textbf{en} is Nigerian English and \textbf{fr} is African French.\label{fig:multilingual}}
\end{figure*}

As a final exploration of the transferability of the representations, we evaluate the representations on a diverse language set of languages with varying amounts of training data and compare the relative reductions in word error rate we obtain when using standard features and switching to the CPC-8k representations. As most of the dataset are small, we trained \textbf{DeepSpeech2 small} models with the Adam optimizer with a fixed learning rate of 0.0002 and applied gradient clipping with maximum norm 25.0, using the same configuration for all languages. Figure \ref{fig:multilingual} summarizes results. Since the experiments above showed that CPC-LibriSpeech features performed badly, we only compare the relative error rediction with CPC-8k features over spectrogram features. In all cases, we find that the CPC-8k representations improve performance relative to spectorgram feature baselines. The largest improvement was obtained on Sundanese where the WER with spectrogram was 27.85 but dropped to 11.49 using CPC-8k features. 

\paragraph{Discussion} As our pre-training data did not have any language labels, it is unclear how many samples were seen for each language during pre-training. However, it is important to know that the \emph{uncurated} multilingual pre-training can improve speech recognition performance on many languages. These results suggests, in practice, that one could use a universal speech feature extractor for many languages instead of training one for each language individually~\cite{kannan2019large}.

\subsection{Control: English Speech Recognition}
Thus far, we have focused on robustness and transferability and seen that CPC-8k features offer considerable benefits in these dimensions compared to traditional features. It remains to demonstrate how well they work in powerful architectures where large amounts of labeled training data is available. To test this, we used 10\% and 100\% portions of LibriSpeech dataset to train speech recognition models, again comparing different features. Our architecture is a standard \textbf{TDNN}. The speech recognition models are trained in the similar way as standard models~\citep{collobert2016wav2letter,openseq2seq}. The models are trained with Adam optimizer with learning rate $0.0002$ and gradient clipping with a maximum norm $5.0$ together with the  polynomial learning rate decay method with power $2.0$ is used over $200$ epochs.\footnote{These hyperparameters were chosen to give optimal performance with baseline log filterbank features, and used, unchanged for our learned features.}

Table \ref{tb:result-tdnn-librispeech} summarizes the results with \textbf{TDNN} models trained on different sizes of LibriSpeech dataset. We see that even if the speech recognition models have a large number of parameters and are trained on plenty of supervised data, the learned representations still provide significant improvements. The pattern continues to hold if we use beam search decoding with a language model.\footnote{\url{http://www.openslr.org/resources/11/4-gram.arpa.gz}} Our \textbf{+ LM decoding} results are comparable to the OpenSeq2Seq benchmark, since we used the exact same LM and decoding algorithm as they used~\citep{openseq2seq}.

Although better results contain be obtained using newer architectures than TDNN~\cite{park2019specaugment,synnaeve2019end}, it still represents a standard and important recognition architecture and the results prove that the representations learned from diverse and noisy data can improve large speech recognition model on English in both low-data and high-data regimes.

\begin{table*}[ht]
    \begin{center}\small
    \begin{tabular}{lrrrrrrrr}
    \toprule
      & \multicolumn{8}{c}{LibriSpeech}\\
      & \multicolumn{2}{c}{dev-clean} & \multicolumn{2}{c}{dev-other} & \multicolumn{2}{c}{test-clean} & \multicolumn{2}{c}{test-other}\\
      & \multicolumn{1}{c}{10\%} & \multicolumn{1}{c}{100\%} &  \multicolumn{1}{c}{10\%} & \multicolumn{1}{c}{100\%} & \multicolumn{1}{c}{10\%} & \multicolumn{1}{c}{100\%} & \multicolumn{1}{c}{10\%} & \multicolumn{1}{c}{100\%}\\
    \midrule
    \multicolumn{9}{l}{\textbf{LibriSpeech}}\\
    LogFilterbank (OpenSeq2Seq) & -& \underline{6.67} & - & \underline{18.67} & - & \underline{6.58} & - & \underline{19.61}\\
    LogFilterbank (ours) & 19.83 & 6.63 & 38.97 & 18.77 & 19.65 & 6.43 & 41.26 & 20.16\\
    CPC-LibriSpeech	& 15.07 & 6.70 & 33.55 & 19.77 & 14.96 & 6.91 & 36.05 & 21.60\\	
    CPC-8k	& \textbf{13.92} & \textbf{6.20} & \textbf{30.85} & \textbf{17.93} & \textbf{13.69} &\textbf{6.25} & \textbf{32.81} & \textbf{19.10}\\
    \midrule
    \multicolumn{9}{l}{\textbf{+ LM decoding}}\\
    LogFilterbank (OpenSeq2Seq) & - & \underline{4.75} & - & \underline{13.87} & - & \underline{4.94} & - & \underline{15.06}\\
    LogFilterbank (ours) & 12.49 & 4.87 & 28.71 & 14.14 & 12.29 & 5.04 & 31.03 & 15.25\\
    CPC-LibriSpeech  & 9.66 & 4.87 & 24.72 & 14.34 & 9.41 & 5.05 & 26.77 & 16.06\\	
    CPC-8k & \textbf{8.86} & \textbf{4.35} & \textbf{22.10} & \textbf{12.96} & \textbf{8.70} & \textbf{4.72} & \textbf{24.15} & \textbf{14.47}\\
    \bottomrule
    \end{tabular}
    \end{center}
    \caption{Sample efficiency experiments with the \textbf{TDNN} trained and evaluated on \textbf{LibriSpeech}. The results are word error rate on the LibriSpeech development and evaluation sets. 10\% vs. 100\% indicates the amount of training data used. The section in \textbf{+ LM decoding} contain results with beamsearch decoding with a 4-gram language model. The underlined (OpenSeq2Seq) scores are taken from public benchmarks.\footnote{\url{https://nvidia.github.io/OpenSeq2Seq/html/speech-recognition/wave2letter.html}}}
    \label{tb:result-tdnn-librispeech}
\end{table*}

\section{Related Work}\label{sec:citation}
Unsupervised learning played an import role in the reintroduction of deep networks to speech processing~\citep{hinton2012deep}, as well as other application areas~\citep{hinton2006fast,bengio2007greedy,vincent2010stacked}. After a period of focusing on supervised techniques, unsupervised representation learning has recently seen a resurgence in a variety of modalities \citep{doersch2017multi,oord2018representation,donahue2019large,bachman2019learning} and has led to improved results, especially in low-data regimes~\citep{henaff2019data,schneider2019wav2vec}. In natural language processing, pretrained representations can outperform state-of-the-art system even in high data regimes~\citep{mikolov2013distributed,devlin2018bert}.

The last two years have produced a large amount of work on unsupervised speech representation learning. Some of this work has been evaluated in terms of its ability to perform phone recognition and similar audio classification tasks~\citep{oord2018representation}. Like us, \citet{schneider2019wav2vec,baevski2019vq} applied learned representations to speech recognition tasks and evaluated on how well in-domain WER was improved. However, as we argued in the paper, such an evaluation misses the opportunity to assess whether these systems become more robust to domain shift and to what extent the learned representations appropriate for different languages.

Finally, the ZeroSpeech challenges have explicitly looked at correlations between learned representations and phonetic structures that generalize across many languages and adapt to new speakers~\cite{dunbar2017zero,dunbar2019zero}. \citet{kahn2019libri} learned representations with contrastive predictive coding on 60,000 hours of English speech and could show that their representations are correlated well with English phonetic structure; however, they did not evaluate these representations in a supervised speech recognizer. 

Recently, there have been considerable improvements in purely supervised speech recognition systems. Data augmentation~\citep{park2019specaugment}, self-training~\citep{synnaeve2019end,kahn2019self} have advanced the state-of-the-art performance on English speech recognition. It is likely that augmentation methods are orthogonal to the proposed improvements on universal speech representation learning, and that one could combine both to improve results even further. Additionally, the impact of data augmentation and self-training can be further assessed in terms of its impact on robustness using the methods proposed in this paper.

\section{Conclusion}\label{sec:conclusion}
We have introduced an unsupervised speech representation learning method that discovers acoustic representations from up to 8000 hours of diverse and noisy speech data.
We have shown, for the first time, that such pretrained representations lead speech recognition systems to be robust to domain shifts compared to standard acoustic representations, and compared to representations trained on smaller and more domain-narrow pretraining datasets. These representations were evaluated on a standard speech recognition setup where the models are trained and evaluated on in-domain data and also on transfer tasks where the models are evaluated on out-of-domain data. We obtained consistent improvements on 25 phonetically diverse languages including tonal and low-resource languages. This suggests we are making progress toward models that implicitly discover phonetic structure from large-scale unlabelled audio signals.

\bibliography{tacl2018}
\bibliographystyle{acl_natbib}

\end{document}